# Omni-Roach: A legged robot capable of traversing multiple types of large obstacles and self-righting

Jonathan Mi, Yaqing Wang, and Chen Li[1]

***Abstract*—Robots excel at avoiding obstacles but struggle to traverse complex 3-D terrain with cluttered large obstacles. By contrast, insects like cockroaches excel at doing so. Recent research in our lab elucidated how locomotor transitions emerge from locomotor-environment interaction for diverse locomotor challenges abstracted from complex 3-D terrain and the strategies to overcome them. Here we built on these fundamental insights to develop a cockroach-inspired legged robot, Omni-Roach, that integrated these strategies to achieve multi-modal locomotion and provide a robophysical model to study the trade-off between multi-functionality and performance. The robot was based on the RHex design with six compliant legs and featured a rounded body with two wings that can open and a tail with pitch and yaw degrees of freedom. After two development and testing iterations, our robot was capable of overcoming all locomotor challenges with a high performance and success rate. It traversed cluttered rigid pillars only 1.1× robot body width apart, a 2.5× hip height bump, a 0.75× body length gap, densely cluttered flexible beams only 65% body width apart, and self-righted within 4 seconds. Systematic beam traversal experiments further revealed that a downward-pointing tail oscillating laterally helps roll the body into beam gaps and break frictional and interlocking contact to traverse. Our work highlights the usefulness of multi-functional appendages and exaptation for large obstacle traversal.**

## I. Introduction

Robots excel at navigating environments with sparse obstacles by creating a geometric map of the environment [1], [2] and planning and following a collision-free trajectory to avoid obstacles [3]–[6]. Yet robots are still poor at traversing complex 3-D terrain cluttered with large obstacles comparable to body size, due to a lack of understanding of locomotor-terrain physical interaction [7]. This has hindered critical applications such as earthquake search and rescue in rubble [8], environmental monitoring on a forest floor [9], and extraterrestrial exploration through Martian rocks [10], [11]. By contrast, insects like the discoid cockroach are excellent at using physical interaction to traverse complex 3-D terrain, often transitioning across different locomotor modes [12].

An important advantage of animals over robots is their multi-functionality, often achieved via exaptation [13], i.e., the ability to co-opt morphology initially evolved for one purpose for another. For example, the discoid cockroach can use the same body and legs to walk [14], run [15], climb [16], and traverse uneven [17] and cluttered [12] obstacles, as well as co-opt wings to self-right [18]. Ghost crabs creatively use the same set of legs in different ways to excavate, pass, and pack sand when burrowing [19]. Trap-jaw ants' powerful mandibles for prey capture are co-opted for jumping [20]. Some robots are already capable of multi-functional locomotion, ranging from millimeter-scale soft-robots that walk, jump, climb, swim, crawl, and grip [21], [22] to large robots transitioning across aerial, aquatic, and terrestrial locomotion [23]–[25]. However, how to use body and appendages to generate multi-modal locomotion to traverse complex 3-D terrain remains poorly understood.

Recent studies in our lab using an interdisciplinary terradynamics approach have begun to reveal how locomotor transitions of insects and legged robots in complex 3-D terrain emerge from physical interaction and how they can be controlled by multi-functional use of body and appendages [7]. This was achieved by systematic animal, robot, and physics studies of model systems of distinct locomotor challenges (Fig. 1b) abstracted from locomotion in complex 3-D terrain (Fig. 1a). For each model system, locomotor-environment interaction leads to stereotyped locomotor modes (Fig. 2, i). Our physics models validated against animal and robot experiments revealed how a suite of design, actuation, and action strategies (Fig. 2, ii) can increase performance for each locomotor challenge by enhancing the transition to desired modes or suppressing undesired ones.

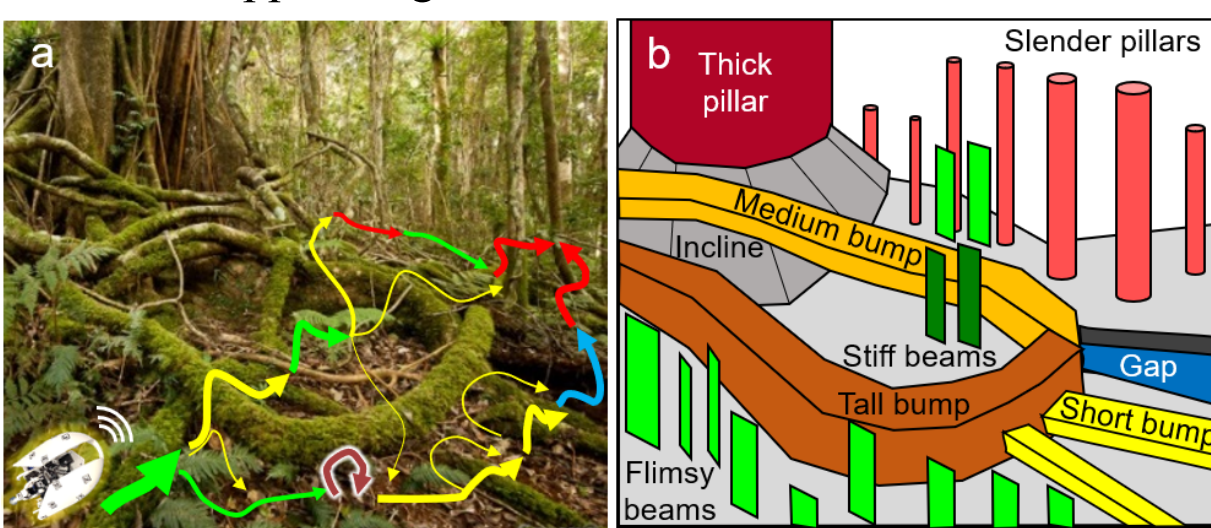

**Fig. 1. Abstracting locomotor transitions in complex 3-D terrain with large obstacles.** (a) Envisioned capability of robot traversing complex 3-D terrain. (b) Abstracted challenges from diverse large obstacles. Reproduced from [7] under terms of CC-BY license.

Building on these foundations, here we take the next step towards robust robot traversal of complex 3-D terrain, by developing and testing a cockroach-inspired legged robot, Omni-Roach, capable of overcoming multiple locomotor challenges using one set of morphology. This robot also provides a robophysical model [26]–[30] for studying the trade-off between multi-functionality and single task performance. Based on insights from each model system, we proposed how to integrate them in a single robot to achieve multi-functionality (Sec. II). We first developed and tested an initial prototype (Sec. III). Based on the limitations revealed, we refined the robot and improved its performance in overcoming multiple locomotor challenges (Sec. IV). We further conducted systematic experiments to understand how to exapted tail use to facilitate traversal of cluttered beam obstacles (Sec. V). Finally, we reviewed our contribution and discussed future directions (Sec. VI). Design files and control code can be found at https://github.com/TerradynamicsLab/Omni-Roach.

[1] Dept. of Mechanical Engineering, Johns Hopkins University. J.M. is with the Department of Electrical & Computer Engineering, University of California, San Diego and performed the research at JHU in an REU program. Corresponding author: [redacted], https://li.me.jhu.edu

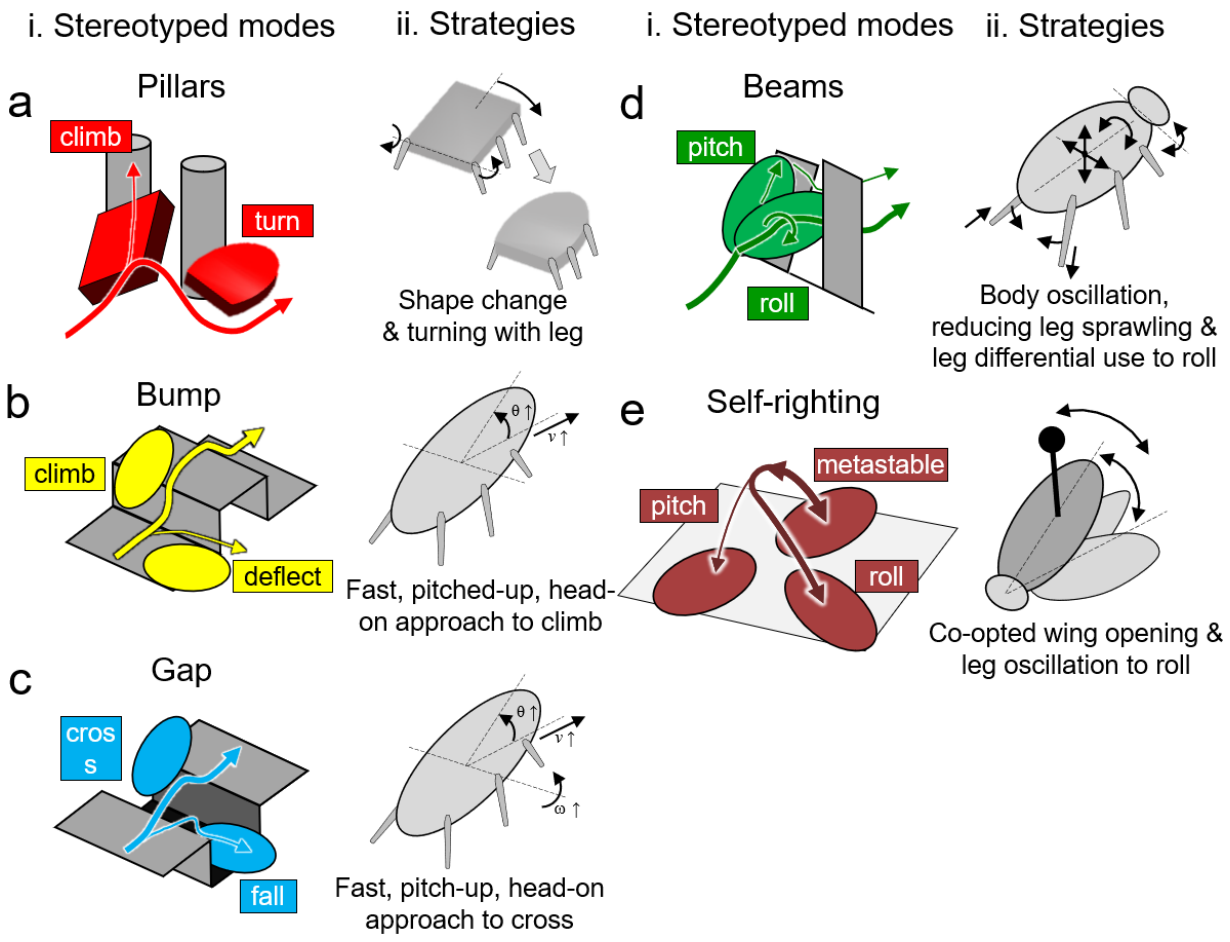

**Fig. 2. Fundamental principles of strategies to overcome abstracted challenges.** (i) Stereotyped modes discovered for each locomotor challenge. (ii) Strategies to modulate locomotor transitions between modes to better overcome challenges. Adapted from [7] under terms of CC-BY license.

## II. Design for Multi-functionality

We first review strategies to overcome the distinct locomotor challenges revealed by our previous studies to inform how to design and control a single robot to overcome these challenges. The robot is based on the classic RHex platform [30], with six legs spinning in an alternating tripod gait.

(i) To better traverse cluttered, rigid, large vertical obstacles such as pillars (Fig. 2a) [31], a robot can adopt an elliptical body shape that passively repels it from obstacles to maintain a desired turn mode. By contrast, the cuboidal body common in robots is attracted to obstacles during self-propulsion and leads to an undesired climb mode.

(ii) To better traverse large horizontal obstacles with height increase like a bump (Fig. 2b) [32], a robot should move towards the bump rapidly, head-on, and pitch the body up to facilitate transitioning into a desired climb mode. An active tail with a pitch degree of freedom can pitch the body up using inertial effect.

(iii) To better traverse large horizontal obstacles with height drop like a gap (Fig. 2c) [33], strategies similar to those for bump traversal facilitate transitioning to a desired cross mode, and an active tail is helpful in a similar fashion.

(iv) To better traverse densely cluttered large, flexible obstacles such as grass-like beams (Fig. 2d), a robot should roll into a gap and maneuver through. This can be facilitated by a rounded body [12], large body oscillations [34], and differential leg use and body flexion [35]. A rounded body helps the robot align with and roll into the gap and reduce the potential energy barrier to move through it [12]. Large body oscillations provide kinetic energy fluctuation to overcome the potential energy barrier [34]. Finally, differential leg use assists body rolling while body flexion breaks resistive frictional and interlocking contact between the body and edge of beams [35].

(v) To self-right after flipping over, a likely scenario when moving through large obstacles [12], [36] (Fig. 2e), simultaneous wing opening and lateral leg oscillations facilitate transitioning from an undesired metastable body orientation to a desired roll mode that results in an upright orientation [18], [37]–[39]. Wing opening lifts the center of mass and reduces the base of support formed by robot-ground contact points, reducing the potential energy barrier to self-right [37], [39], [40]. Lateral leg oscillations generate kinetic energy fluctuation that perturbs the body to overcome the barrier [39], [40].

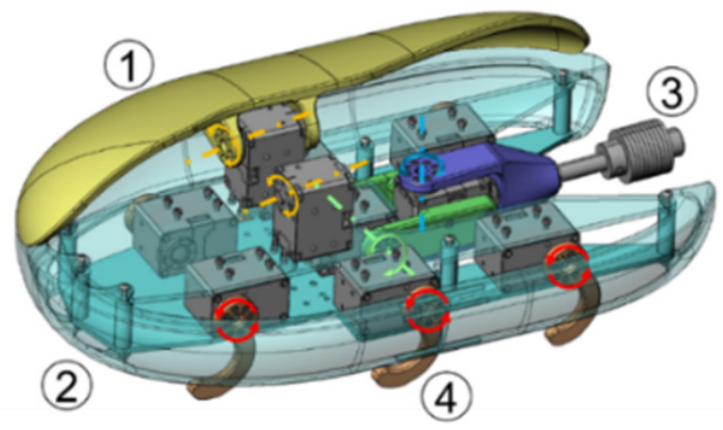

**Fig. 3. Omni-Roach v1 CAD model.** (1) Actuated wings that can open (left wing is not shown). (2) Fixed bottom shell. (3) Active 2-DOF tail. (4) Compliant C-shaped legs. Arrows show rotational joints, red for leg joints, yellow for wing joints, and green and blue for tail pitch and yaw joints, respectively.

Based on these insights, we identified three design features that can be integrated into a single robot for overcoming these locomotor challenges (multimedia material, movie 1). (1) A rounded body to help passively deflect the robot from rigid vertical obstacles and roll into gaps to traverse cluttered flexible obstacles. (2) An active tail to pitch the body up for traversing large horizontal obstacles and to generate body oscillations for traversing cluttered flexible obstacles and self-righting. (3) The ability to open its top shell as wings for self-righting.

## III. Initial Robot Development & Testing

### A. Design

The initial prototype, Omni-Roach v1 (Fig. 3), measures 35 cm long (excluding the tail), 23 cm wide, 13 cm tall, and weighs 1.8 kg, with a 5 cm hip height. We designed the legs to be protected under the shell so that they do not get caught easily on obstacles, and the shell should be as compact as possible in width to reduce roll inertia, and then in length. The chassis plates were laser cut from acrylic. The six compliant C-shaped legs were 3D-printed from elastic thermoplastic polyurethane (TPU). Other custom components were 3D-printed from polylactic acid (PLA). Besides six servo motors that spin the legs (Fig. 3, red arrows), two motors actuate a tail with both pitch and yaw degrees of freedom (Fig. 3, green and blue arrows). The tail weighs 0.17 kg (9% of total body mass) and measures 19 cm (54% body length) from the pitch pivot axis to the tip. It is defined to be at home position (pitch = yaw = 0°) when the tail points straight backward (position shown in Fig. 3). The rounded body shell was divided into the top and bottom halves. The top is separated into the left and right halves, each capable of opening via rolling by a servo motor (Fig. 3, yellow arrows). We used DYNAMIXEL XL430-W250-T servo motors (no-load speed 0.95 rev/s, stall torque 1.4 N·m) for all joints. The robot was controlled by a Python program. The motors are set to operate at their maximum speed.

### B. Testing

We built simple testbeds to study the robot for each locomotor challenge (Fig. 4a-c). Each obstacle was mounted to a pegboard as the ground. The pillar (Fig. 4a) was an aluminum bar vertically fixed to the ground. The bump obstacle (Fig. 4b) was stacked square aluminum bars. Sandpaper (60 grit) on the ground was used to improve leg traction. The beams were acrylic plates (6 mm thick) mounted on the ground via a joint with torsional springs (Fig. 4c). We chose a high torsional stiffness of $K$ = 0.78 N·m/rad so that the robot could not simply push over the beams to traverse.

The robot was manually controlled via keyboard in all obstacle traversal experiments. Whenever the legs spun, they followed an alternating tripod gait.

(i) For pillar traversal, the robot was driven straight towards the pillar with no other control.

(ii) For bump traversal, the robot initially lifted its tail up. When the head contacted the bump, the tail pitched down to the home position to pitch the body up using inertial effects. Once the head reached above the bump, the tail was further lowered to raise the rest of the body so the robot could use legs to climb onto the bump completely.

(iii) For gap traversal, the robot initially lifted its tail up. When the head was near the gap edge (various distance was tested), the tail pitched down to the home position to pitch the body up using inertial effects.

(iv) For beam traversal, the robot was driven straight towards the beam gap. As the robot pitched up against the beams and oscillated within the gap, the tail pitched down, resulting in body rolling. Based on real-time observation of the body yaw direction, the robot was commanded (by experience) to oscillate the tail while keeping the legs spinning to align the head with the gap.

For self-righting, the robot first opened the wings to raise the body, and the tail pitched downward relative to the body and yawed to one side (Fig. 5d, i). After the robot rolled onto its side, the wings closed, and the tail moved back to the home position (Fig. 5d, ii). The wing contacting the ground further pushed the robot upright (Fig. 5d, iii).

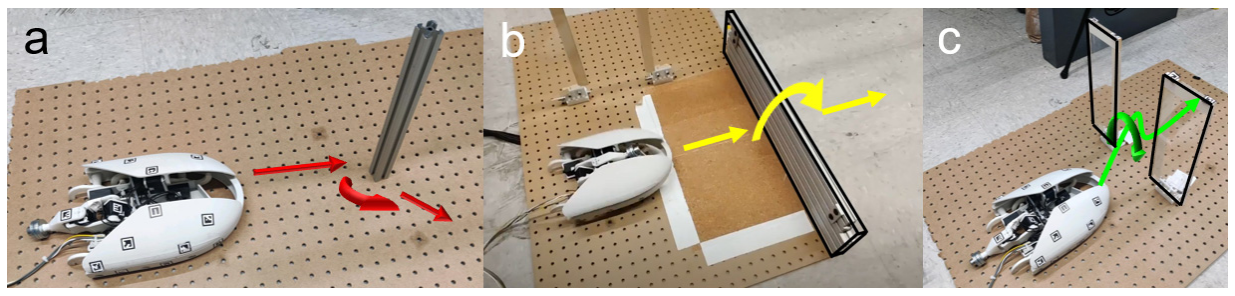

**Fig. 4. Simple terrain testbeds.** (a) Pillar. (b) Bump. (c) Beams. Colored arrows show favored locomotor modes, with colors following Fig. 2.

### C. *Performance*

#### *1) Pillar traversal*

The rounded body enabled Omni-Roach v1 to deflect away and traverse pillars (Fig. 5a) 100% of the time ($N$ = 10) (multimedia material, movie 2).

#### *2) Bump traversal*

Without tail use, Omni-Roach v1 could not pitch its body sufficiently by simply pushing against the bump and was deflected away (Fig. 5b, iii). With tail use, the robot pitched up to climb over the bump (Fig. 5b, i-ii) before it deflected. By cutting the rear end of the shell (Fig. 5b, i, red circle) to reduce chassis-ground contact, the robot traversed a bump of up to 1.5× hip height 80% of the time ($N$ = 10) (multimedia material, movie 3).

#### *3) Gap traversal*

Omni-Roach v1 traversed a 17 cm gap (48% body length) without tail actuation. This was not improved with tail actuation ($N$ = 10), because the slow spinning leg motors limited the robot speed to below the necessary speed for dynamic gap traversal. The slow tail motor also did not generate sufficient body pitching via inertial effects.

#### *4) Beam traversal*

The active tail was designed for two purposes: to pitch the body up using inertial effects for bump and gap traversal, and to generate lateral perturbation for self-righting. In beam traversal testing we found that it can also be used to tap the ground to facilitate traversal. Without tail use, Omni-Roach v1 pitched up against the beams and became stuck during self-propulsion, unable to traverse (Fig. 5c, iii). By keeping the tail pitched downward and manually yawing it left and right at roughly 2 Hz, the tail pushed against the ground and helped the body roll, which, together with intermittent leg propulsion against the beams, enabled the robot to traverse (Fig. 5c, i-ii). Similar to how cockroaches adjust their legs and abdomen to traverse the beams [35] and a recent robot [41], the robot's tail tapping and intermittent leg propulsion worked together to oscillate the robot and break the frictional and interlocking contact with the beams (multimedia material, movie 4).

By varying beam gap width, we found that Omni-Roach v1 could traverse beams with a gap 91% of the robot body width (or a 21 cm gap, Fig. 9c). To quantify how much the body rolled to achieve this, we estimated the approximate maximal body roll. Neglecting body thickness:

$$\text{body roll} = \cos^{-1}\frac{\text{beam gap}}{\text{body width}} \tag{1}$$

To traverse the 91% body width gap, the robot rolled by approximately 24°.

#### *5) Self-righting*

Omni-Roach v1 can self-right robustly with a 90% success rate and within 4 s on flat ground ($N$ = 10) (multimedia material, movie 5).

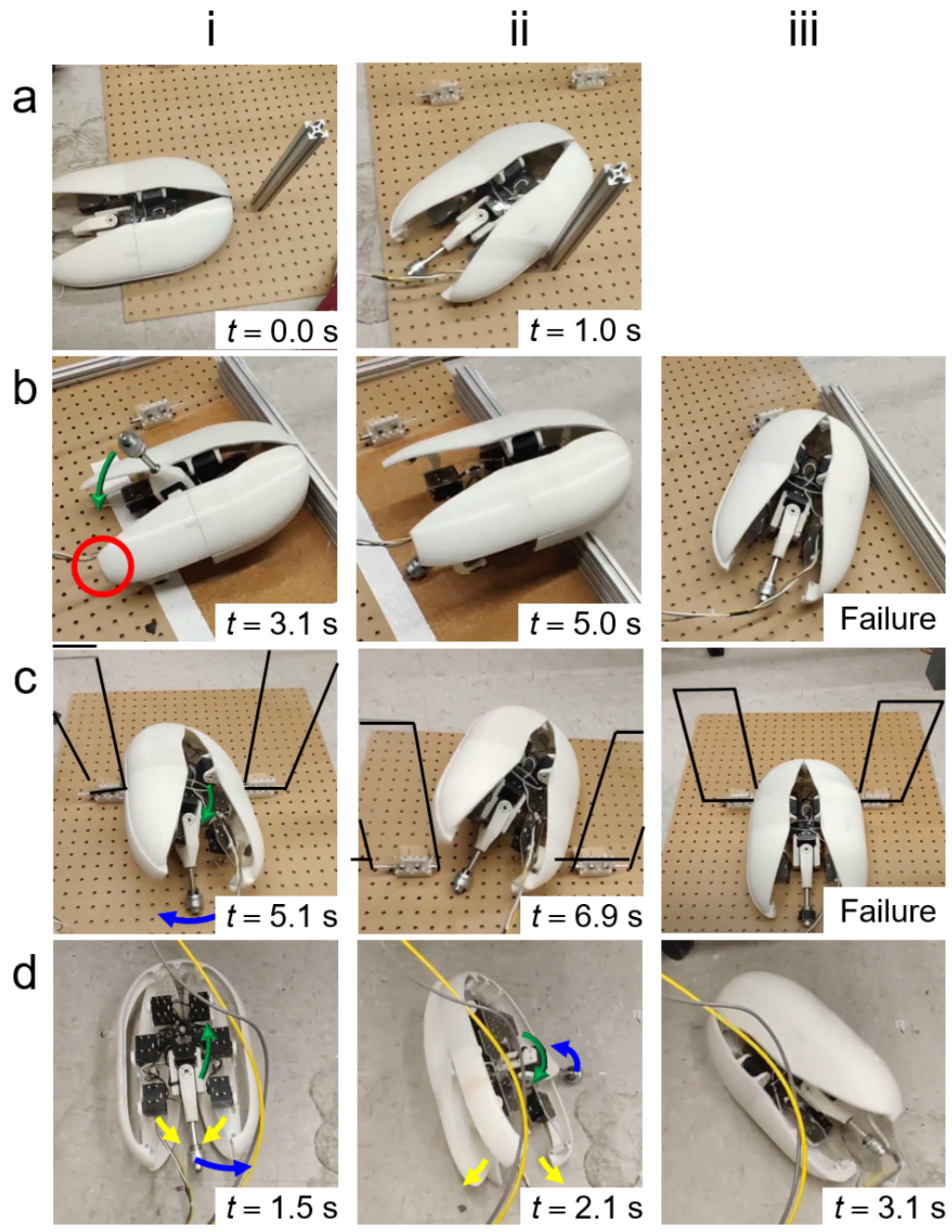

**Fig. 5. Snapshots of Omni-Roach v1 overcoming locomotor challenges.** (a) Pillar. (b) Bump of 1.5× hip height. Red circle shows the clipped-off wings and bottom chassis. (c) Beams with a 91% body width gap. (d) Self-righting. Blue and green arrows show tail yaw and pitch motions. Yellow arrows show wing motions.

### D. Limitations revealed

Omni-Roach v1 was capable of completing the bump, pillar, beam, and self-right challenges. However, the performance of the robot on the bump, beam, and gap could be improved. We speculated that further reducing the size and weight of the robot could make it more agile. Tail usage was important for bump and beam traversal, so improvements to the tail could be impactful. Improvements to the legs could also enable better forward propulsion.

## IV. Refined Robot Development & Testing

### A. Design

Learning from the results of Omni-Roach v1, we developed Omni-Roach v2 (Fig. 6a) to improve performance. Omni-Roach v2 measures 20 cm long (excluding the tail), 18.5 cm wide, 10 cm tall, and weighs 0.75 kg. Its hip height is the same as Omni-Roach v1 at 5 cm. Overall, Omni-Roach v2 is about 2/3 the size of v1 (Fig. 6b) and less than half the weight, making it much more agile. Despite the shortened and flatter chassis, the leg length of v2 remains the same as v1, giving the legs more reach relative to body size. The tail weighs 0.1 kg (13% of total body mass) and measures 14.5 cm (73% body length) from the pitch pivot axis to the tip. The tail home position is defined the same way as in v1 (position shown in Fig. 6a). The tail was lengthened relative to body length to reduce chassis-ground contact, which hindered bump traversal. We used the smaller, lighter, but faster DYNAMIXEL XL330-M288-T servo motors (no-load speed 1.71 rev/s, stall torque 0.52 N·m). The legs were also changed from C- to S-shape to double stride frequency. The motors operated at maximum speed in all experiments.

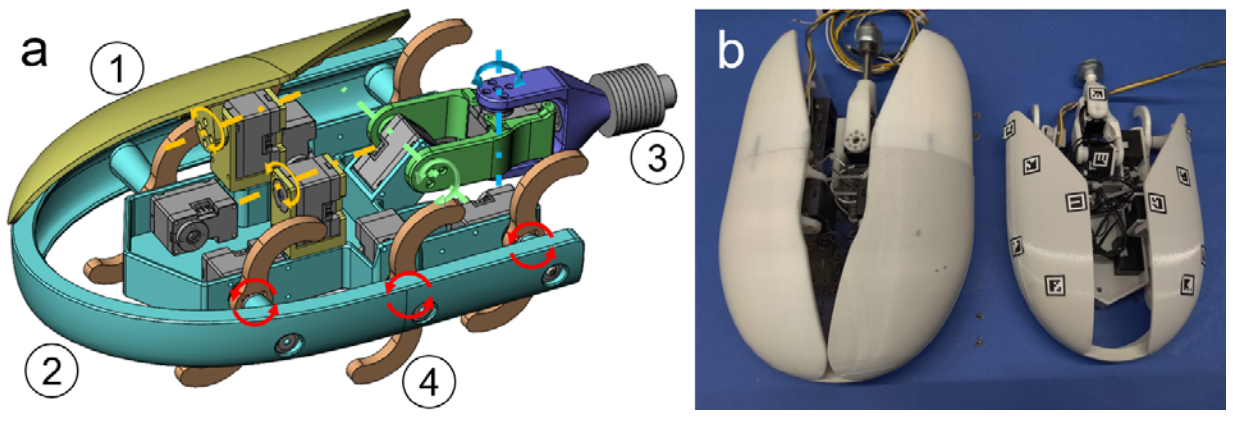

**Fig. 6. Omni-Roach v2.** (a) CAD model. (1) Actuated wings that can open (left wing is not shown). (2) Fixed bottom shell. (3) Active 2-DOF tail. (4) Compliant S-shaped legs. Arrows show rotational joints, red for leg joints, yellow for wing joints, and green and blue for tail pitch and yaw joints, respectively. (b) Size comparison between v1 (left) and v2 (right).

### B. Multi-obstacle testbed

Besides testing for individual types of obstacles like for v1, to demonstrate Omni-Roach v2's ability to traverse complex terrain with multiple types of challenges, we created a multi-obstacle field (Fig. 7) consisting of four pillars spaced 20 cm, a bump obstacle, and a pair of beams. The robot traversed all the obstacles, was manually flipped over, and self-righted (multimedia material, movie 7). We used the same traversal strategy and control logic as v1.

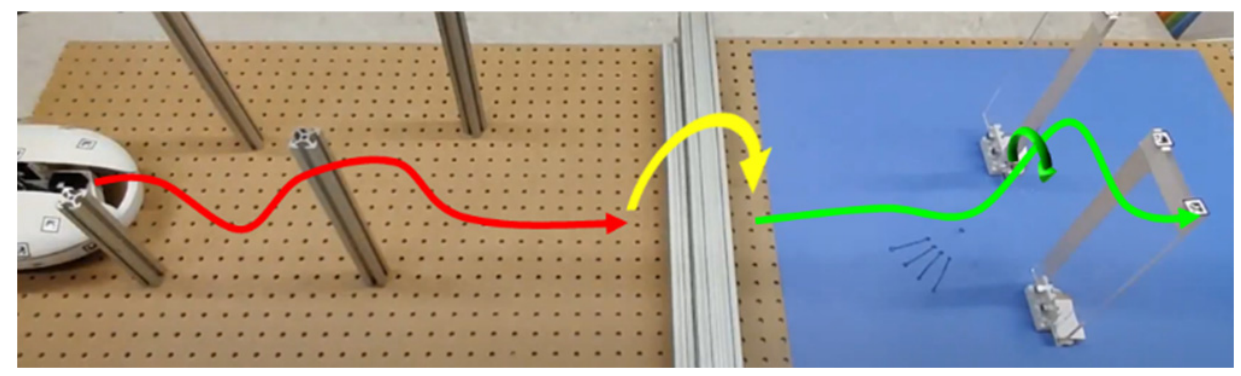

**Fig. 7. Multi-obstacle field with four pillars, a bump, and two beams.** Arrows show favored locomotor modes, with colors following Fig. 2.

### C. Performance

#### 1) Pillar traversal

The rounded body enabled Omni-Roach v2 to deflect away from the pillar (Fig. 8a) 100% of the time ($N$ = 10) (multimedia material, movie 2).

#### 2) Bump traversal

With tail use, Omni-Roach v2 pitched up to climb over a maximal bump height of 2.5× hip height (Fig. 8b, i-ii) with an 80% success rate ($N$ = 10) (multimedia material, movie 3).

#### 3) Gap traversal

Omni-Roach v2 traversed a 13.5 cm gap (68% body length) with the tail stowed in the body ($N$ = 10) and a 15 cm gap (75% body length) with the tail in the home position to help bridge the gap ($N$ = 10). Despite Omni-Roach v2 being smaller, lighter, and faster, the performance was not improved by involving a tail motion ($N$ = 10). The leg and tail motors are too slow to generate sufficient forward speed and sufficient body pitching to dynamically traverse as in [33].

#### 4) Beam traversal

Omni-Roach v2 utilizes the same techniques as Omni-Roach v1 to traverse the beams. Omni-Roach v2 could traverse beams (Fig. 8c, i-ii) with a gap of only 65% of the robot body width (or a 12 cm, Fig. 9d). This translates to approximately a 50° body roll (multimedia material, movie 4). Beam traversal of Omni-Roach v2 is further discussed in Sec. V.

#### 5) Self-right performance

Omni-Roach v2 self-righted (Fig. 8d, i-iii) using the same motion sequence as Omni-Roach v1, taking only 4.0 s with a 100% success rate on flat ground ($N$ = 10) (multimedia material, movie 5).

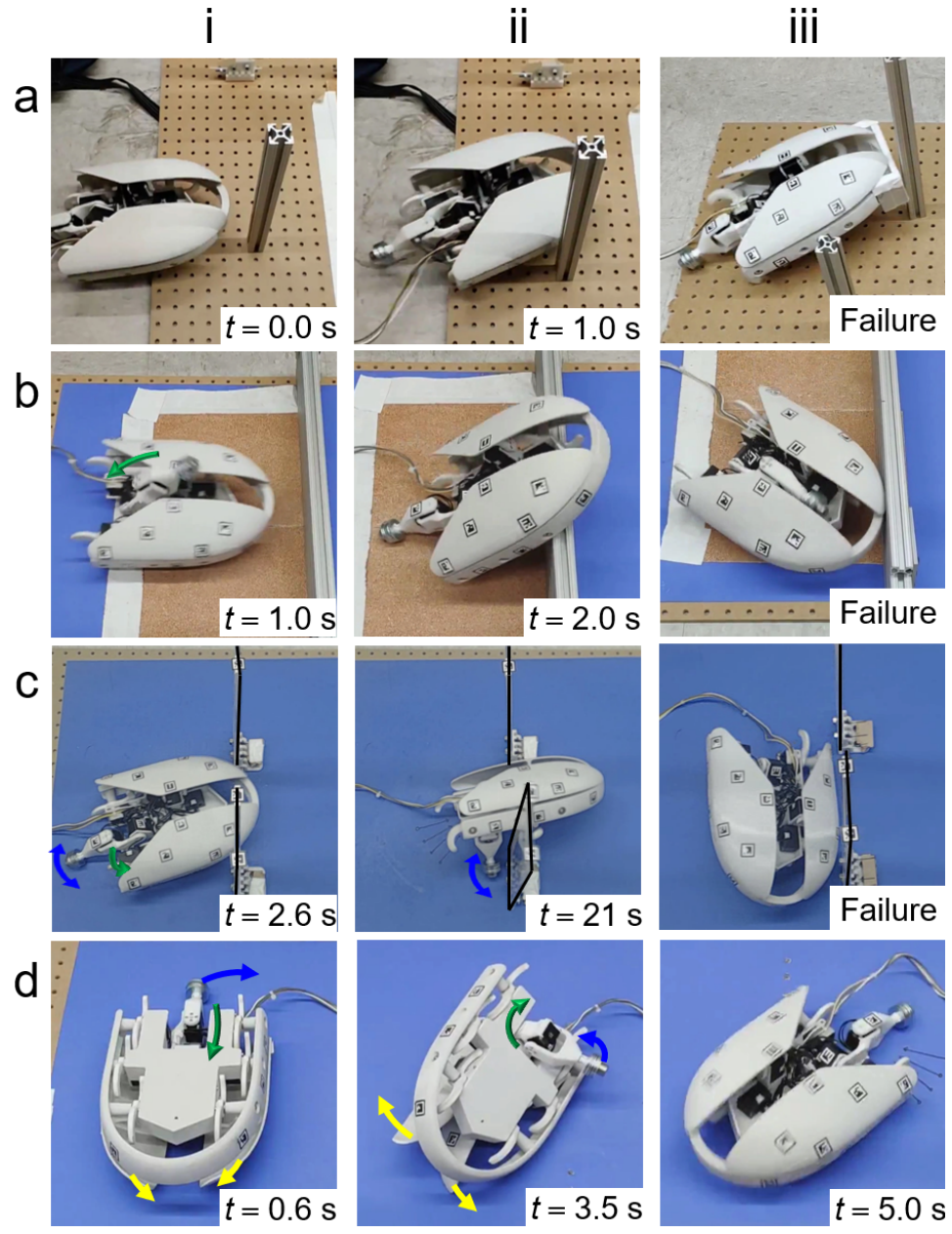

**Fig. 8. Snapshots of Omni-Roach v2 overcoming locomotor challenges.** (a) Pillar. (b) Bump of 2.5× hip height. Red circle shows the clipped-off wings and bottom chassis. (c) Beams with a 65% body width gap. (d) Self-righting. Blue and green arrows show tail yaw and pitch motions. Yellow arrows show wing motions.

### D. Comparison to v1

A comparison of size and performance between Omni-Roach v1 and v2 is listed in Tab. 1. Omni-Roach v2 had a higher performance in bump and gap traversal, capable of traversing a bump of 2.5× hip height compared to 1.5× for v1 (Fig. 9a-b) and a gap of 75% body length compared to 48% for v1. It also had a better performance in beam traversal, capable of rolling into the beams by 50° to traverse a 65% body width gap, compared to rolling by 24° to traverse a 91% body width gap for v1 (Fig. 9c-d). It maintained the same level of performance in pillar traversal and self-righting.

This performance increase was mainly due to the increase in the relative leg length (from 14% to 25% body length) and relative tail length (from 54% to 73% body length), while maintaining relative leg propulsion (single leg stall force/body weight only slightly reduced from 1.6 to 1.4). In addition, Omni-Roach v2 ran faster than v1 (0.30 vs. 0.26 m/s) and had a stride frequency about three times that of v1 (2.7 vs. 0.9 Hz), which generated more oscillations during beam interaction and better perturbed the body to break frictional and interlocking contact.

Omni-Roach v2 showed no worse performance than v1.

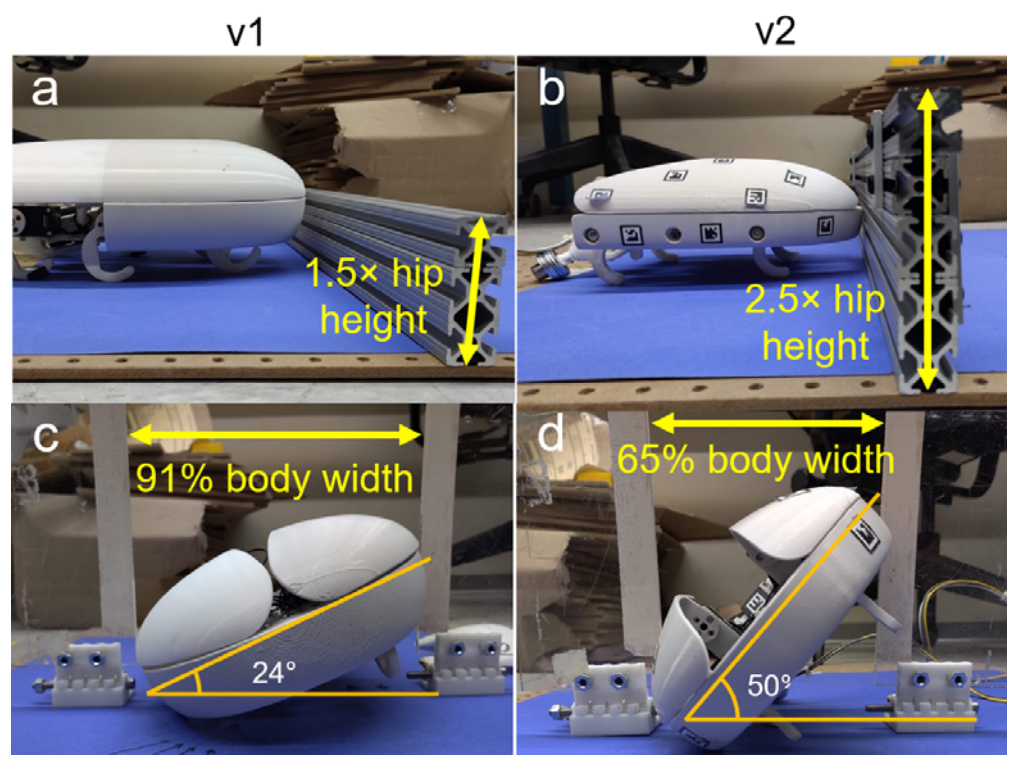

**Fig. 9. Performance comparison between v1 and v2.** (a-b) Bump of maximal height traversed. (c-d) Beams of minimal beam gap width traversed, showing maximal body roll achieved.

TABLE I
COMPARISON BETWEEN OMNI-ROACH V1 AND V2

| | v1 | v2 |
|---|---|---|
| Length (cm) | 35 | 20 |
| Width (cm) | 23 | 18.5 |
| Height (cm) | 13 | 10 |
| Weight (kg) | 1.8 | 0.75 |
| Hip height (cm) | 5 | 5 |
| Pillar traversal probability | 100% | 100% |
| Maximum bump height (× hip height) | 1.5 | 2.5 |
| Maximum gap length (× body length) | 48% | 75% |
| Roll angle to traverse beams (deg) | 24 | 50 |
| Self-right time (sec) | 4 | 4 |

## V. TAIL USE FOR BEAM TRAVERSAL

Among these challenges, we observed that cluttered beam traversal required the most finesse of coordination between tail pitch and yaw, taking the longest to complete. In addition, a recent study of a RHex-class robot traversing uneven terrain discovered that, by introducing downward tail tapping to perturb the body, the robot could be freed from being stuck in uneven terrain [41]. We speculated that similar active tail tapping can perturb our robot from being stuck when pushing against the beams by breaking frictional and interlocking contact. Aside from tail use, during testing, we also noticed that the robot would become stuck in the beams at some approach angles but not others. Considering these, we further studied how tail use and approach angle affect beam traversal of Omni-Roach v2.

### *A. Experimental setup and procedure*

In this experiment, we used a beam gap of 65% robot body width (or 12 cm). We tested four tail operation modes. (1) Home static: a static tail with 0° pitch and 0° yaw relative to the chassis (Fig. 10a). (2) Pitch down static: a static tail pointing downward, with 90° pitch and 0° yaw relative to the chassis (Fig. 10b). (3) Pitch down and yaw static: a static tail pointing downward and sideways, with 90° pitch and 15° yaw relative to the chassis (Fig. 10c). (4) Pitch down with yaw oscillation: a downward pointing, laterally oscillating tail, with 90° pitch relative to the chassis, and a manually controlled yaw oscillation from −15° to 15° at 2 Hz (Fig. 10d). This mode was inspired by a beam traversal experiment with cockroaches [35]. When the animal first rolled into the gap, it flexed its abdomen down (abdomen pitched from 7° to 37° relative to the body) and oscillated the abdomen to propel itself forward. In (4) pitch down with yaw oscillation tail mode, the tail pitch resembled the abdomen pitch and the yaw oscillation resembled the abdomen oscillation in the animal experiment. The other modes were chosen to compare with (4).

We varied approach angle, the angle between the midline of the two beams to the robot's midline (Fig. 10e), from 0° to 45° in increments of 15°. With other conditions kept the same, we collected 10 trials for each combination of tail mode and approach angle, resulting in a total of 160 trials (multimedia material, movie 6).

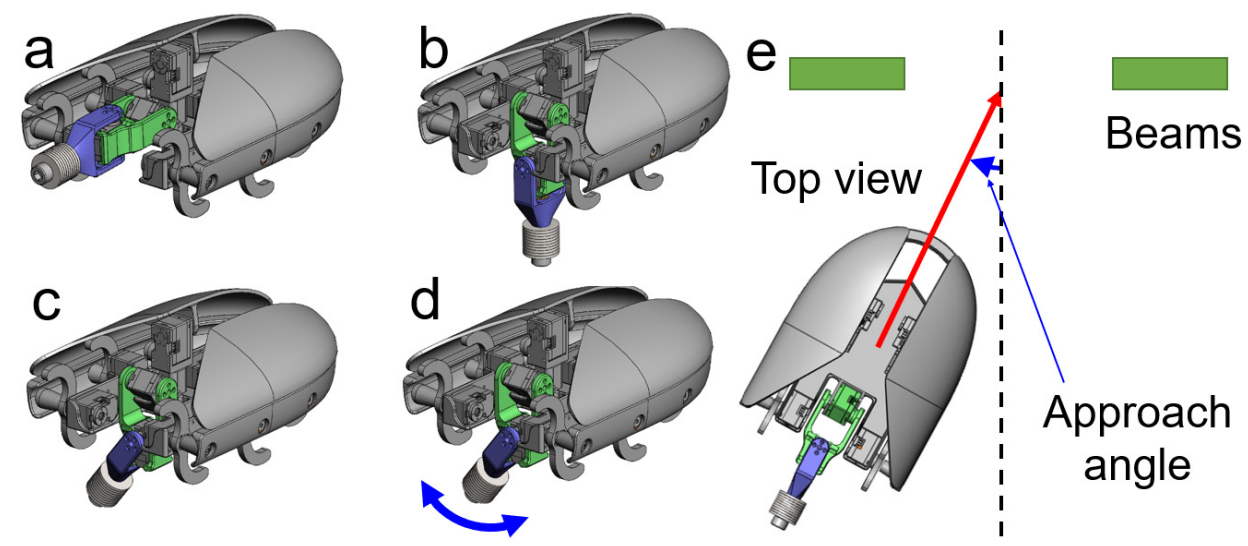

**Fig. 10. Experimental design to study the effects of tail use and approach angle for beam traversal.** Four tail modes tested: (a) Home static. (b) Pitch down static. (c) Pitch down and yaw static. (d) Pitch down with yaw oscillation. Arrows show yaw oscillation. (e) Definition of approach angle.

### *B. Results*

Traversal probability depended sensitively on tail operation mode (Fig. 11a). In the home static tail mode, Omni-Roach v2 could not traverse at any approach angle because the small roll perturbation from leg propulsion was insufficient to roll the body into the gap. Instead, the robot pitched up and oscillated against the beams and became trapped.

In the pitch down and yaw static tail mode, the robot was also unable to traverse at any approach angle. Although the statically yawed tail rolled the robot body, the resistance between the body and beams prevented the robot from fully entering the gap. As a result, the robot was deflected away 75% of the time and trapped between the beams 25% of the time.

In the pitch down static tail mode, the robot was unable to traverse at a 0° approach angle but did so with a high probability at the 15°, 30°, or 45° approach angles. At 0° approach angle, the robot head could not fully enter the gap, which caused it to become stuck or deflected away. At larger approach angles, the head entered the gap, which led to body rolling that enabled traversal.

In the pitch down with yaw oscillation tail mode, the robot was able to traverse at all four approach angles with a high probability. The oscillating tail tapped against the ground, which served two functions. First, it rolled the body substantially and helped it enter the gap. Second, it generated perturbations to break frictional and interlocking contact to free the body from being stuck when maneuvering through the gap. As a result, traversal time (Fig. 11a) of the pitch down with yaw oscillation tail mode was reduced by an average of 30% from that of the pitch down static tail mode at 15°, 30°, or 45° approach angles ($P < 0.001$, ANOVA).

We further compared average traversal time between different tail modes where the robot traversed at a high probability, defined as the time from when the robot first touched the beams to when the robot fully left the beams after traversal through the gap. (Fig. 11b). The robot traversed more quickly using the pitch down with yaw oscillation tail mode than the pitch down static tail mode at all approach angles ($P < 0.001$, repeated-measures ANOVA). Traversal time also reduced with approach angle ($P < 0.01$, ANOVA).

We noted that, without tail oscillation, a static tail that pitched downward to support the body could allow the robot to roll into the gap and traverse, but only when the approach angle is appropriate. Thus, we speculated that one function of tail oscillation is to adjust the approach angle.

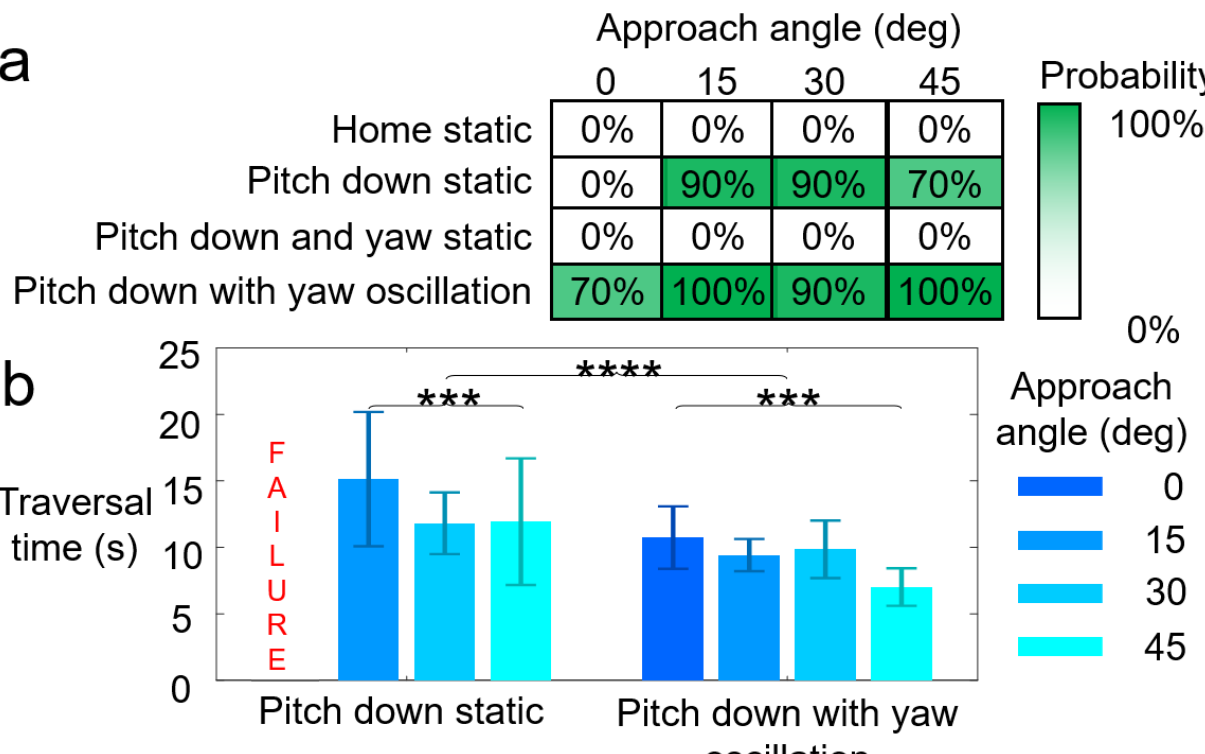

**Fig. 11. Beam traversal performance of Omni-Roach v2.** (a) Traversal probability. (b) Average traversal time of successful trials for different approach angles for pitch down static (left) and pitch down with yaw oscillation (right) tail modes. ***: $P < 0.01$, ****: $P < 0.001$, ANOVA.

## VI. Summary & Future Work

We developed and tested a small, multi-functional legged robot that can traverse multiple types of large obstacles and self-right, capabilities required for effective operation in complex 3-D terrain. Through two development and testing iterations, our robot can traverse a field of cluttered rigid pillars with 1.08× body width gaps via passive interaction, traverse a large bump of 2.5× its hip height, a large gap of 0.75× its body length gap, and densely cluttered beams with gaps of 65% its body width with active tail use as an exaptation, and self-right within 4.0 s co-opting rounded shell as active wings along with tail exaptation. Systematic experiments showed that the robot traversed beams with a high probability with appropriate tail use.

Many multi-legged robots can walk or run over modest terrain with small height changes relative to leg length over each step by maintaining a stable upright body posture [42]–[45]. Our approach enabled robots to destabilize the body to generate dynamic locomotor transitions to traverse cluttered obstacles comparable to body size. This significantly expands the accessible terrain and save payloads for applications like search and rescue and extraterrestrial exploration [11].

Our work opens opportunities for several further studies. First, we can improve the robot actuators and leg design to enable dynamic gap traversal and improve bump and beam traversal. With smaller and faster motors (as in [32], [33]) and improved leg geometry (like minitour [46] or GOAT [47] legs), we can let the robot generate a higher speed approaching a gap or bump. Increasing the tail inertia relative to the robot driven by a sufficiently powerful motor can enhance body pitching. These will increase gap and bump traversal performance [32], [33]. Higher degree-of-freedom legs can also improve beam traversal by allowing adjustments to generate active rolling torque and reduce leg resistance against obstacles [35].

In addition, we need to enable autonomous locomotor transitions for the robot. Feedback control [48] using the sensed information about terrain interaction is likely crucial for achieving this. For example, the discoid cockroach adjusts its legs and flex the head and abdomen to transition between locomotor modes to better maneuver through densely cluttered beam obstacles [35]. It is likely that this involves sensory feedback using contact and force information gathered by mechanoreceptors, including exteroceptors like bristles that directly sense exterior forces and proprioceptors like hair plates, campaniform sensilla, and chordotonal organs that sense internal displacements and force to infer external forces [49], [50]. Similarly, by closing the loop with contact force sensing [51]–[53] to detect physical properties of large obstacles [40] to complement terrain geometry information from vision, we can develop algorithms to control the robot to change actuation patterns to switch modes autonomously.

Finally, we can use our robot as a physical model to study the trade-off between single-task performance and multi-functionality in animal locomotion. During evolution, as an appendage or a body part of an animal becomes more specialized for a specific task or goal, it loses multi-functionality in others [54], [55]. However, the trade-off between these two aspects is often difficult to study in animals due to difficulty in modifying traits without changing behavior [26]. Instead, robophysical models allow systematically variation of relevant design parameters while controlling others to quantify and understand the trade-off and discover optimal solutions and insights on robot design [56]–[58].

## Acknowledgment

J.M. thanks Qihan Xuan and Ratan Othayoth for discussion and the Laboratory for Computational Sensing & Robotics (LCSR) of the Johns Hopkins University for hosting his REU. We thank Xiao Yu for naming the robot. This work was supported by an Arnold & Mabel Beckman Foundation Beckman Young Investigator Award and a Burroughs Wellcome Fund Career Award at the Scientific Interface to C.L. and an NSF research experience for undergraduates (REU) award hosted by JHU LCSR to J.M. Author contributions: J.M. designed robots, performed experiments, and wrote the paper; Y.W. analyzed data, provided feedback on robot design and experiment, and revised the paper; C.L. oversaw the study and revised the paper.

## REFERENCES

[1] M. W. M. G. Dissanayake et al., 'A solution to the simultaneous localization and map building (SLAM) problem', *IEEE Trans. Robot. Autom.*, vol. 17, no. 3, pp. 229–241, Jun. 2001.

[2] A. Elfes, 'Using occupancy grids for mobile robot perception and navigation', *Computer (Long. Beach. Calif).*, vol. 22, no. 6, pp. 46–57, Jun. 1989.

[3] J. Borenstein and Y. Koren, 'The vector field histogram-fast obstacle avoidance for mobile robots', *IEEE Trans. Robot. Autom.*, vol. 7, no. 3, pp. 278–288, Jun. 1991.

[4] O. Khatib, 'Real-time obstacle avoidance for manipulators and mobile robots', in *Proceedings. 1985 IEEE International Conference on Robotics and Automation*, 1986, vol. 2, pp. 500–505.

[5] E. Rimon and D. E. Koditschek, 'Exact robot navigation using artificial potential functions', *IEEE Trans. Robot. Autom.*, vol. 8, no. 5, pp. 501–518, 1992.

[6] S. Thrun, 'Toward robotic cars', *Commun. ACM*, vol. 53, no. 4, pp. 99–106, Apr. 2010.

[7] R. Othayoth et al., 'Locomotor transitions in the potential energy landscape-dominated regime', *Proc. R. Soc. B Biol. Sci.*, vol. 288, no. 1949, p. rspb.2020.2734, Apr. 2021.

[8] F. Matsuno et al., *Field and Service Robotics*. London: Springer London, 1998.

[9] G. Freitas et al., 'Design, Modeling, and Control of a Wheel-Legged Locomotion System for the Environmental Hybrid Robot', in *Biomechanics / 752: Robotics*, 2011, pp. 302–310.

[10] L. Matthies et al., 'Computer Vision on Mars', *Int. J. Comput. Vis.*, vol. 75, no. 1, pp. 67–92, Jul. 2007.

[11] C. Li and K. Lewis, 'The Need for and Feasibility of Alternative Ground Robots to Traverse Sandy and Rocky Extraterrestrial Terrain', *Adv. Intell. Syst.*, no. 2100195, Jan. 2022.

[12] C. Li et al., 'Terradynamically streamlined shapes in animals and robots enhance traversability through densely cluttered terrain', *Bioinspiration and Biomimetics*, vol. 10, no. 4, p. 46003, 2015.

[13] S. J. Gould and E. S. Vrba, 'Exaptation—a Missing Term in the Science of Form', *Paleobiology*, vol. 8, no. 1, pp. 4–15, Feb. 1982.

[14] C. P. Spirito and D. L. Mushrush, 'Interlimb Coordination During Slow Walking in the Cockroach: I. Effects of Substrate Alterations', *J. Exp. Biol.*, vol. 78, no. 1, pp. 233–243, Feb. 1979.

[15] R. Kram et al., 'Three-dimensional kinematics and limb kinetic energy of running cockroaches.', *J. Exp. Biol.*, vol. 200, no. 13, pp. 1919–1929, Jul. 1997.

[16] R. E. Ritzmann et al., 'Descending control of body attitude in the cockroach Blaberus discoidalis and its role in incline climbing', *J. Comp. Physiol. A*, vol. 191, no. 3, pp. 253–264, Mar. 2005.

[17] S. Sponberg and R. J. Full, 'Neuromechanical response of musculo-skeletal structures in cockroaches during rapid running on rough terrain', *J. Exp. Biol.*, vol. 211, no. 3, pp. 433–446, Feb. 2008.

[18] C. Li et al., 'Cockroaches use diverse strategies to self-right on the ground', *J. Exp. Biol.*, vol. 222, no. 15, Aug. 2019.

[19] D. Springthorpe, 'Biomechanical Multifunctionality in the Ghost Crab, Ocypode quadrata', Ph.D. dissertation, Integrative Biology, University of California, Berkeley, Berkeley, CA, 2016.

[20] S. N. Patek et al., 'Multifunctionality and mechanical origins: Ballistic jaw propulsion in trap-jaw ants', *Proc. Natl. Acad. Sci.*, vol. 103, no. 34, pp. 12787–12792, Aug. 2006.

[21] W. Hu et al., 'Small-scale soft-bodied robot with multimodal locomotion', *Nature*, vol. 554, no. 7690, pp. 81–85, Feb. 2018.

[22] E. B. Joyee et al., '3D Printed Biomimetic Soft Robot with Multimodal Locomotion and Multifunctionality', *Soft Robot.*, vol. 9, no. 1, pp. 1–13, Feb. 2022.

[23] R. J. Lock et al., 'Multi-modal locomotion: from animal to application', *Bioinspir. Biomim.*, vol. 9, no. 1, p. 011001, Dec. 2013.

[24] S. Mintchev and D. Floreano, 'Adaptive Morphology: A Design Principle for Multimodal and Multifunctional Robots', *IEEE Robot. Autom. Mag.*, vol. 23, no. 3, pp. 42–54, Sep. 2016.

[25] K. H. Low et al., 'Perspectives on biologically inspired hybrid and multi-modal locomotion', *Bioinspir. Biomim.*, vol. 10, no. 2, p. 020301, Mar. 2015.

[26] J. Aguilar et al., 'A review on locomotion robophysics: the study of movement at the intersection of robotics, soft matter and dynamical systems', *Reports Prog. Phys.*, vol. 79, no. 11, p. 110001, Nov. 2016.

[27] M. H. Dickinson et al., 'Wing Rotation and the Aerodynamic Basis of Insect Flight', *Science*, vol. 284, no. 5422, pp. 1954–1960, Jun. 1999.

[28] G. V. Lauder et al., 'Fish biorobotics: kinematics and hydrodynamics of self-propulsion', *J. Exp. Biol.*, vol. 210, no. 16, pp. 2767–2780, Aug. 2007.

[29] A. V. Birn-Jeffery et al., 'Don't break a leg: running birds from quail to ostrich prioritise leg safety and economy on uneven terrain', *J. Exp. Biol.*, vol. 217, no. 21, pp. 3786–3796, Nov. 2014.

[30] U. Saranli et al., 'RHex: A Simple and Highly Mobile Hexapod Robot', *Int. J. Rob. Res.*, vol. 20, no. 7, pp. 616–631, Jul. 2001.

[31] Y. Han et al., 'Shape-induced obstacle attraction and repulsion during dynamic locomotion', *Int. J. Rob. Res.*, vol. 40, no. 6–7, pp. 939–955, Jun. 2021.

[32] S. W. Gart and C. Li, 'Body-terrain interaction affects large bump traversal of insects and legged robots', *Bioinspir. Biomim.*, vol. 13, no. 2, p. 026005, Feb. 2018.

[33] S. W. Gart et al., 'Dynamic traversal of large gaps by insects and legged robots reveals a template', *Bioinspir. Biomim.*, vol. 13, no. 2, p. 026006, Feb. 2018.

[34] R. Othayoth et al., 'An energy landscape approach to locomotor transitions in complex 3D terrain', *Proc. Natl. Acad. Sci.*, vol. 117, no. 26, pp. 14987–14995, Jun. 2020.

[35] Y. Wang et al., 'Cockroaches adjust body and appendages to traverse cluttered large obstacles', *bioRxiv:2021.10.02.462900*, 2021.

[36] C. Li et al., 'Mechanical principles of dynamic terrestrial self-righting using wings', *Adv. Robot.*, vol. 31, no. 17, pp. 881–900, Sep. 2017.

[37] Q. Xuan and C. Li, 'Randomness in appendage coordination facilitates strenuous ground self-righting', *Bioinspir. Biomim.*, vol. 15, no. 6, p. 065004, Nov. 2020.

[38] Q. Xuan and C. Li, 'Coordinated Appendages Accumulate More Energy to Self-Right on the Ground', *IEEE Robot. Autom. Lett.*, vol. 5, no. 4, pp. 6137–6144, Oct. 2020.

[39] R. Othayoth and C. Li, 'Propelling and perturbing appendages together facilitate strenuous ground self-righting', *eLife*, vol. 10, pp. 1–23, Jul. 2021.

[40] Q. Xuan et al., 'Environmental force sensing enables robots to traverse cluttered obstacles with interaction', *arXiv:2112.07900*, 2021.

[41] D. Soto et al., 'Enhancing Legged Robot Navigation of Rough Terrain via Tail Tapping', in *Robotics for Sustainable Future*, Springer, Cham, 2022, pp. 213–225.

[42] J. Tan et al., 'Sim-to-Real: Learning Agile Locomotion For Quadruped Robots', in *Robotics: Science and Systems XIV*, 2018.

[43] J. Hwangbo et al., 'Learning agile and dynamic motor skills for legged robots', *Sci. Robot.*, vol. 4, no. 26, pp. 1–14, Jan. 2019.

[44] D. Wooden et al., 'Autonomous navigation for BigDog', in *2010 IEEE International Conference on Robotics and Automation*, 2010, pp. 4736–4741.

[45] A. Bouman et al., 'Autonomous Spot: Long-Range Autonomous Exploration of Extreme Environments with Legged Locomotion', in *2020 IEEE/RSJ International Conference on Intelligent Robots and Systems (IROS)*, 2020, pp. 2518–2525.

[46] G. Kenneally et al., 'Design Principles for a Family of Direct-Drive Legged Robots', *IEEE Robot. Autom. Lett.*, vol. 1, no. 2, pp. 900–907, Jul. 2016.

[47] S. Kalouche, 'Design for 3d agility and virtual compliance using proprioceptive force control in dynamic legged robots', Ph.D. dissertation, School of Computer Science, Carnegie Mellon University, Pittsburgh, PA, 2016.

[48] M. H. Dickinson et al., 'How Animals Move: An Integrative View', *Science*, vol. 288, no. 5463, pp. 100–106, Apr. 2000.

[49] J. C. Tuthill and R. I. Wilson, 'Mechanosensation and Adaptive Motor Control in Insects', *Curr. Biol.*, vol. 26, no. 20, pp. R1022–R1038, Oct. 2016.

[50] J. C. Tuthill and E. Azim, ‘Proprioception’, *Curr. Biol.*, vol. 28, no. 5, pp. R194–R203, Mar. 2018.

[51] X. A. Wu et al., ‘Tactile Sensing and Terrain-Based Gait Control for Small Legged Robots’, *IEEE Trans. Robot.*, vol. 36, no. 1, pp. 15–27, Feb. 2020.

[52] J. J. Shill et al., ‘Terrain identification on a one-legged hopping robot using high-resolution pressure images’, in *2014 IEEE International Conference on Robotics and Automation (ICRA)*, 2014, pp. 4723–4728.

[53] Y. Wang et al., ‘A robophysical model to study physical sensing of obstacles in legged traversal of complex 3-D terrain’, in *Bulletin of the American Physical Society 66*, 2021, vol. S14, no. 007.

[54] K. A. Corn et al., ‘A Multifunction Trade-Off has Contrasting Effects on the Evolution of Form and Function’, *Syst. Biol.*, vol. 70, no. 4, pp. 681–693, Jun. 2021.

[55] A. Pirrone et al., ‘When natural selection should optimize speed-accuracy trade-offs’, *Front. Neurosci.*, vol. 8, no. 8 APR, pp. 1–5, Apr. 2014.

[56] N. Gravish and G. V. Lauder, ‘Robotics-inspired biology’, *J. Exp. Biol.*, vol. 221, no. 7, pp. 1–8, Apr. 2018.

[57] A. J. Ijspeert et al., ‘From Swimming to Walking with a Salamander Robot Driven by a Spinal Cord Model’, *Science*, vol. 315, no. 5817, pp. 1416–1420, Mar. 2007.

[58] J. T. Schultz et al., ‘Using a biologically mimicking climbing robot to explore the performance landscape of climbing in lizards’, *Proc. R. Soc. B Biol. Sci.*, vol. 288, no. 1947, p. rspb.2020.2576, Mar. 2021.